\documentclass[journal]{IEEEtran}

\usepackage{amssymb}
\usepackage{amsmath}
\usepackage{ifpdf}

\usepackage{cite}

\ifCLASSINFOpdf
   \usepackage[pdftex]{graphicx}
   \graphicspath{{../pic/}}
   \DeclareGraphicsExtensions{.txt,.jpeg,.png}
\else
\fi
\usepackage{hyperref}
\usepackage{amsmath}
\usepackage{graphicx}
\usepackage{float} 
\usepackage{booktabs} 
\usepackage{subfigure}
\usepackage{array}
\usepackage{fixltx2e}

\usepackage{stfloats}
\usepackage{url}
\usepackage{color}

\begin{document}
%
\title{MAP-Net: Multiple Attending Path Neural Network \\ for Building Footprint Extraction from Remote \\ Sensed Imagery}
%
%
%

\author{Qing~Zhu,
        Cheng~Liao,
        Han~Hu,
        Xiaoming~Mei,
        Haifeng~Li\^*,~\IEEEmembership{Member,~IEEE}
\thanks{This work was supported by the National Natural Science Foundation of China (41631174)}
\thanks{Qing~Zhu, Cheng~Liao and Han~Hu are with the Faculty of Geosciences and Environmental Engineering, Southwest Jiaotong University, Chengdu 611756, China (e-mail: zhuq66@263.net; liaocheng@my.swjtu.edu.cn; han.hu@swjtu.edu.cn). }
\thanks{Xiaoming~Mei and Haifeng~Li are with the School of Geosciences and Info-Physics, Central South University, Changsha, 410083, China (e-mail: meixiaoming17@163.com; lihaifeng@csu.edu.cn).}
\thanks{Haifeng~Li is the corresponding author.}
}

%
%

\markboth{to appear in ieee transactions on geoscience and remote sensing, 2019 }%
{Shell \MakeLowercase{\textit{et al.}}: MAP-Net: Multiple Attending Path Neural Network for Building Footprint Extraction from Remote Sensed Imagery}
%



\maketitle

\begin{abstract}
Building footprint extraction is a basic task in the fields of mapping, image understanding and computer vision, etc. Accurately and efficiently extracting building footprints from a wide range of remote sensed imagery remains a challenge due to the complex structures, variety of scales and diverse appearances of buildings. Existing convolutional neural network (CNN)-based building extraction methods are criticized for their inability to detect tiny buildings because the spatial information of CNN feature maps is lost during repeated pooling operations of the CNN. Additionally, large buildings still have inaccurate segmentation edges. Moreover, features extracted by a CNN are always partially restricted by the size of the receptive field,  and large-scale buildings with low texture are always discontinuous and holey when extracted. To alleviate these problems, multi-scale strategies are introduced in the latest researches to extract buildings with different scales. While the features with higher resolution generally extracted from shallow layers, which extracted insufficient semantic information for tinny buildings. This paper proposes a novel multiple attending path neural network (MAP-Net) for accurately extracting multi-scale building footprints and precise boundaries. Unlike existing multi-scale feature extraction strategies, MAP-Net learns spatial localization-preserved multi-scale features through a multi-parallel path in which each stage is gradually generated to extract high-level semantic features with fixed resolution. Then, an attention module adaptively squeezes channel-wise features extracted from each path for optimized multi-scale fusion, and a pyramid spatial pooling module captures global dependency for refining discontinuous building footprints. Experimental results show that our method achieved 0.88\%, 0.93\% and 0.45\% F1-score and 1.53\%, 1.50\% and 0.82\% intersection over union (IoU) score improvements without increasing computational complexity compared with the latest HRNetv2 on the Urban 3D, Deep Globe and WHU datasets, respectively. Specifically, MAP-Net outperforms MA-FCN, which is the state-of-the-art (SOTA) algorithms with post-processing and model voting strategies, on WHU dataset without pre-training and post-processing. The TensorFlow implementation is available at https://github.com/lehaifeng/MAPNet.
\end{abstract}

\begin{IEEEkeywords}
Building footprint extraction, deep learning, semantic segmentation, attention mechanism, remote sensing imagery.
\end{IEEEkeywords}

%
\IEEEpeerreviewmaketitle

\section{Introduction}
%
%
%
%
\IEEEPARstart{T}{he} rapid development of remote sensing technology has made it easier to acquire many high-resolution optical remote sensing images that support the extraction of building footprints in a wide range\cite{bittner2018building,wu2018automatic,liu2019building}. Immediate and accurate building footprint information is significant for illegal building monitoring, 3D building reconstruction, urban planning and disaster emergency response. Due to low inter-class variance and high intra-class variance in buildings in optical remote sensed imagery\cite{kang2019eu}, parking lots, roads and other non-buildings are highly similar to buildings in appearance. With the variety of building materials, scales and illumination, the representation of buildings in remote sensed imagery shows significant differences. Therefore, how to accurately and efficiently extract building footprints from remote sensed imagery remains a challenge.
\par Over the past two decades, numerous algorithms have been proposed to extract building footprints. They can be divided into two categories: traditional image processing-based and CNN-based methods.
\par Traditional building extraction methods utilize the characteristics of the spectrum, texture, geometry and shadow\cite{du2019novel,li2014extracting,gavankar2018automatic,burochin2014detecting,cote2012automatic,zhou2014seamless} to design feature operators for extracting buildings from optical images. Since these features vary under different illumination conditions, sensor types and building architectures, traditional methods can resolve only specific issues for specific data. \cite{awrangjeb2013automatic,du2017automatic,gilani2016automatic,sohn2007data} combined optical imagery with GIS \cite{simonetto2005rectangular}, and digital surface models (DSM) were obtained from light detection and ranging (Lidar) or synthetic aperture radar interferometry \cite{Zhu2017Deep} to distinguish non-building areas that are highly similar to buildings, which increased the robustness of building extraction, although a wide range of corresponding multi-source data acquisition is always costly.
\par Because buildings in remote sensing images are diverse in structure, appearance and scale, building extraction algorithms have evolved from handcrafted feature-based methods to learning feature-based methods, such as deep convolutional neural networks (DCNNs). Moreover, deep networks have been practical in designing CNN models. For instance, ResNet \cite{he2016identity} with 152 layers introduced identity mapping in a residual block to solve the problem of gradient explosion in propagation, making it possible to design a deeper network to extract richer semantic features.
\par Evolving from CNN, fully convolutional network (FCN)-based \cite{Zhu2017Deep},\cite{huang2016building,kemker2018algorithms,sun2018developing,maltezos2018building,deng2018multi,maggiori2016convolutional,bittner2017building,sun2019fusion,alshehhi2017simultaneous,zhang2018fusion} building extraction methods achieve incredible results and are often used in building semantic segmentation tasks. The encoder-decoder framework \cite{bittner2017building},\cite{li2019semantic,ji2019scale,liu2019building,ronneberger2015u,badrinarayanan2017segnet,khalel2018automatic},\cite{hasan2019u} obtains more accurate building footprints than FCN-based methods, particularly on the localization of the boundary since they recover spatial details through skip connections to fuse shallow high-resolution features in the decoder stage. Deep Encoding Network \cite{liu2019net} (DE-Net) introduced lately techniques based on encoder-decoder network for building extraction and achieved higher performance. Nevertheless, coarse features introduced by shallow layers become the main obstacle for accurate building boundaries. \cite{ji2019scale,liu2019building,chen2017deeplab,lin2019esfnet} used a conditional random field (CRF) for post processing to refine the edge of footprints, which achieved great improvements on building boundaries. Multi-scale aggregation FCN \cite{wei2019toward} (MA-FCN) extracted multi-scale features based on feature pyramid network and applied polygon regularization methods for boundary refinement which achieved SOTA result on WHU dataset.
\par For the problem of multi-scale building extraction, \cite{sun2019fusion}, \cite{li2018multiple,ji2019scale} integrated hierarchical results extracted from multiple models or a design-specific CNN architecture to address multi-scale input for accurately extracting multi-scale buildings. CU-Net \cite{wu2018automatic} applied multi-constraints based on FCN to enhance multi-scale feature representation and performed better in building segmentation than FCN. SiU-Net \cite{ji2018fully} designed two weight-shared branches for multi-scale input and was reported to improve the segmentation accuracy especially for large buildings, yet obviously increased the computational complexity. SRI-Net \cite{liu2019building} aggregated multi-scale features which generated from modified ResNet101 by a spatial residual inception module, which significantly retaining local details due to the higher resolution features throughout the network. \cite{zhao2017pyramid} proposed a pyramid spatial pooling module by introducing several global pooling layers to capture multi-scale features without significantly increasing the computational complexity. It is more efficient than \cite{sun2019fusion} and \cite{ji2019scale} in multi-scale building extraction and improves continuously since the global dependency is captured by pooling layers. EU-Net \cite{kang2019eu} proposed a deep spatial pyramids pooling (DSPP) module for multi-scale feature extraction which benefits for extracting localization details and multi-scale buildings. Besides, it introduced focal loss to reduce the impact of mislabelling promoting the training more stability.
\par The attention mechanism \cite{he2019adaptive,fu2019dual,yuan2018ocnet,cao2019gcnet,wang2018non,chen20182,hu2018squeeze} is another method for capturing global relations with long-range dependencies in spatial or channel aspects, which effectively improves the performance of segmentation. Recently, HRNetv1 \cite{sun2019deep} and HRNetv2 \cite{sun2019high} proposed a high-resolution CNN to address multi-scale feature extraction and achieved a new goal in semantic segmentation. While the features with different resolutions were fused each other during the feature extraction, the fine boundary details of building localization information may be lost.
\par In previous studies, CNN-based building footprint extraction algorithms have mainly been encoder-decoder-based, which loses the spatial details in the encoder stage and recovers by fusing shallow feature maps during the decoder stage. However, it causes inaccurate localization on building boundaries since the coarse features introduced from shallow layers and small buildings may be unrecognized. Additionally, the extracted features are always partially restricted by the local respective field, and large-scale buildings with low texture are always discontinuous and holey when extracted.
\par Although many studies have noted the importance of multi-scale features for accurate building extraction, there are still many aspects that need to be improved. This research proposes a MAP-Net inspired by HRNetv2 \cite{sun2019high} to solve the problems described above. First, a parallel multi-path network is generated gradually in each stage which consist of serial convolution blocks to extract high-level semantics and preserve multi-level localization details through serial convolution blocks with fixed spatial resolution in each path. Then, the attention mechanism-based feature enhancement module is introduced to adaptively squeeze channel-wise feature maps from each path for multis-cale feature optimal combination. Pyramid spatial pooling operations follow to extract global semantic information for continuous building footprints. The main contributions of this study are as follows:
\begin{enumerate}
\item We propose a MAP-Net for efficient and accurate multi-scale building footprint boundary extraction through parallel localization-preserved convolutional networks. Different from SOTA methods like HRNetv2, our basic hypothesis is the independence and singularity of path features to be more beneficial. Hence, our strategy is features extracted from each path were independent with fixed scales, and multi-scale features were only fused at the end of the encoder by attention module.
\item We introduce a channel-wise attention module to adaptively squeeze multi-scale features extracted from the multi-path network. These features strengthen the building representation by optimally combining global semantic and spatial localization.
\item We validate the effectiveness of the introduced modules in MAP-Net and high-resolution features extracted from shallow layers on building extraction through extensive ablation studies.
\item The proposed method achieved 0.88\%, 0.93\% and 0.45\% F1-score and 1.53\%, 1.50\% and 0.82\% intersection over union (IoU) score improvements compared with the latest HRNetv2 \cite{sun2019high} on the Urban 3D \cite{goldberg2018urban}, Deep Globe \cite{demir2018deepglobe} and WHU \cite{ji2018fully} datasets and outperforms other SOTA methods on WHU dataset without pre-training.
\end{enumerate}

\par The rest of the paper is organized as follows. Section II introduces the detailed structure of the proposed network for building extraction. Section III describes the experiments and analyses the results. The discussions and conclusions of this paper are presented in Section IV.

\section{Methodologe}
\subsection{Overview}

\begin{figure}[!htb]
\setlength{\abovecaptionskip}{-3pt}
\centering
\includegraphics[scale=0.48]{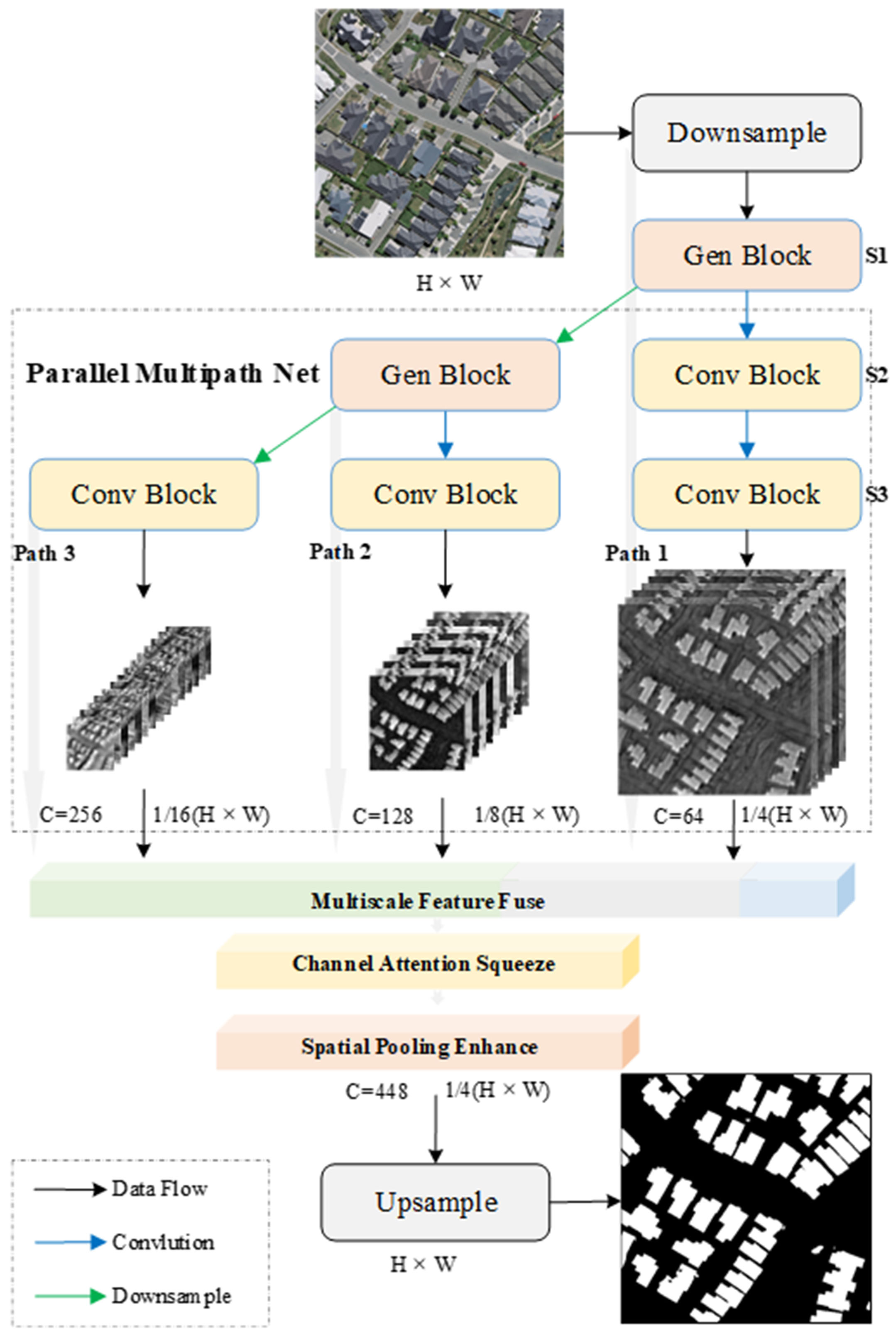} 
\caption{Structure of the proposed MAP-Net, which is composed of three modules: (a) Detail-preserved multi-path feature extraction network;(b) Attention-based multi-scale features adaptive squeeze and global spatial pooling enhancement module; (c) Upsampling and building footprint extraction module. The conv block is composed of a series of residual modules to extract features. A gen block generates a new parallel path to extract higher level semantic features on the basis of conv block.}
\label{Fig.f1_main}
\end{figure}

\par Repeated pooling layers or stride convolution lose spatial localization during the feature extraction procedure. Existing CNN-based building extraction methods recover spatial localization through skip connections to fuse shallow feature maps or upsample feature maps extracted from the last layer by interpolation. However, shallow feature maps contain coarse semantics, introducing noise information in building extraction. In addition, convolutions process the local neighbourhood information and hardly to capture global dependency for large buildings. We propose a MAP-Net for multi-scale building footprint extraction with accurate boundaries and continuous entities. Fig. \ref{Fig.f1_main} illustrates the structure of the proposed MAP-Net. It mainly includes three components: 

\begin{enumerate}
\item A parallel multi-path network to extract multi-scale high-level semantic features while preserving spatial detail information through fixed feature resolution in every path;
\item An attention-based multi-scale features adaptive squeeze and spatial pooling module for global semantic enhancement;
\item An interpolation-based upsampling module for building footprint extraction 
\end{enumerate}

\par As shown in Fig. \ref{Fig.f1_main}, the detail-preserved multi-scale feature extraction network includes three stages, and the parallel path is generated gradually in each stage to extract richer high-level semantic representations with fixed feature spatial resolution to preserve local details. At the end of the feature extract module, multi-scale features extracted from each path were upsampled to the same resolution as features extracted from path 1 by bilinear interpolation and concatenated fusion for attention-based adaptive optimization. The spatial pooling module extracts global dependency to suppress the holes and obtain continuous building entities in the final extraction module. The numbers of channels C and resolution of the feature maps are marked in the figure. H and W represent the height and width of the input image, respectively.

\par The remainder of this section is arranged as follows. Section B presents a detail-preserved multi-path feature extraction network. Attention-based multi-scale features adaptively squeeze, and spatial global pooling enhancement is described in section C. Finally, section D describes the basic unit and training strategies involved in this study.

\subsection{Localization-Preserved Multi-path Network}
\begin{figure}[b]
\setlength{\abovecaptionskip}{-3pt}
\centering
\includegraphics[scale=0.23]{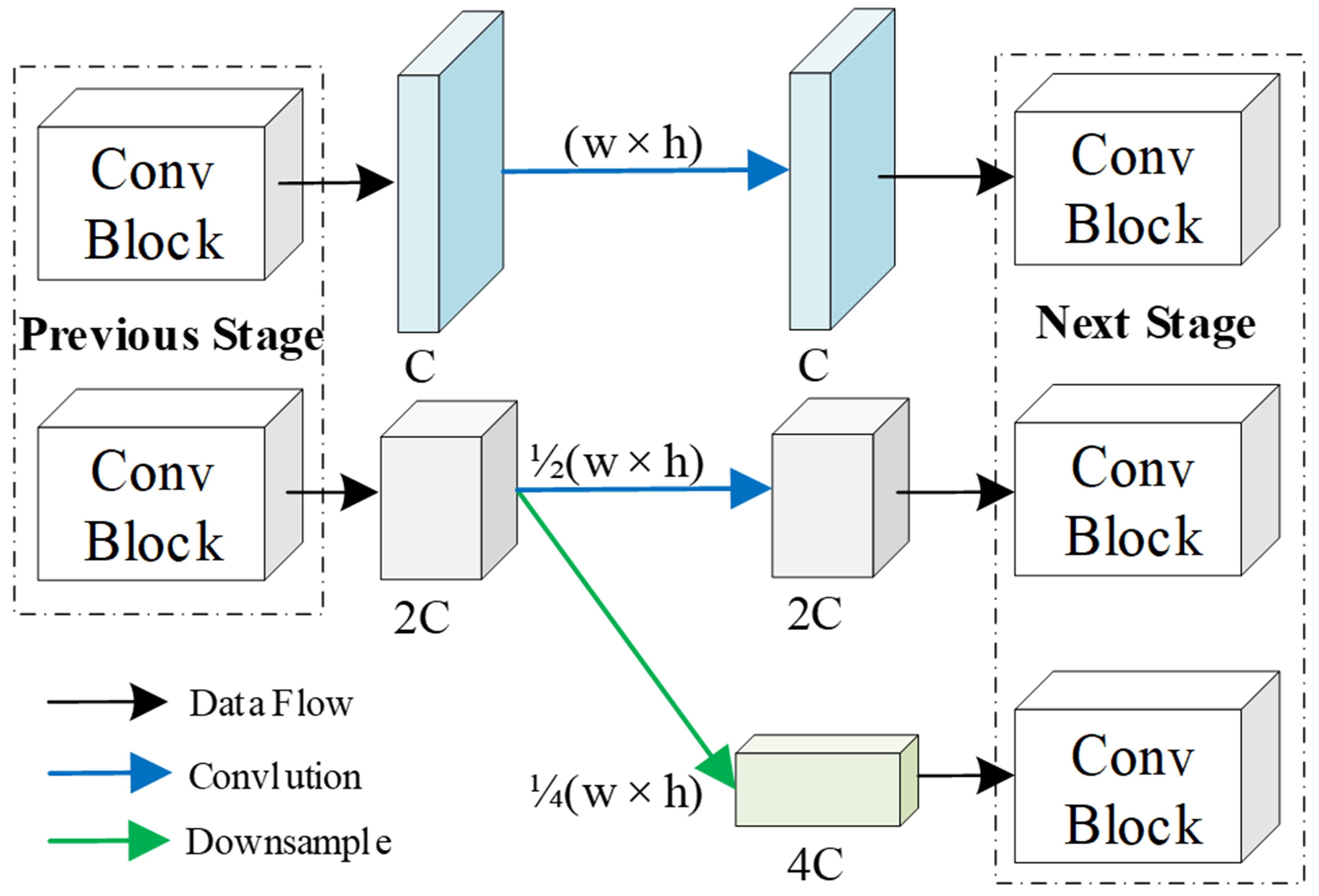} 
\caption{Part of the detail-preserved multi-path network, w and h represent the with and height of the features, respectively. There are two parallel paths with multi-scale features extracted in the previous stage. A new path is generated in the next stage to extract higher-level semantic representations with downsampled resolution and doubled channels. Features in parallel paths are not fused between two stages}
\label{Fig.f2_multipath}
\end{figure}

\par Compared with the encoder-decoder-based CNN structure, the advantage of a localization-preserved multi-path feature extraction network is that it extracts multi-scale features that contain rich high-level semantic representation and accurate spatial localization information rather than recovering spatial details by fusing shallow feature maps during the decoding. Multi-scale features extracted from different stages are fed into several parallel paths that are gradually generated to extract richer semantics and preserve spatial resolution without increasing the computational complexity of the network
\par Fig. \ref{Fig.f2_multipath} illustrates part of the proposed multi-path network. There are two parallel-path extracted feature maps with different dimensions in the previous stage, and a new path is generated in the next stage to extract higher-level semantic features with double-downsampled resolution and double channels as the green arrow shows. It’s composed of a max pooling layer to downsample resolutions and a 1$\times$1 convolutional layer to increase feature channels. The blue arrow represents a 3$\times$3 convolutional layer which transport features from the last convolution block to the next one. The convolution block was composed of four residual blocks to extract high-level semantic representation, details were described in part D. Feature maps extracted in each path maintain spatial resolution, and richer semantic was extracted as the depth of convolution layers increase. Features with higher resolution preserved as many localization details and lower one captured more global semantic. The most important difference from HRNetv2 is that multi-scale features extracted from different paths don’t fuse each other during extraction since the fusion operation may weaken localization details.

\par In the entire process of feature extraction, spatial resolutions and channels of feature maps in each path are fixed. Features in each path are extracted by a series of convolutional blocks that suppress the coarse semantics in high-resolution feature maps compared with encoder-decoder-based CNN. Because detailed representation is preserved in higher-resolution features, smaller buildings and localization of the boundary can be extracted exactly. The effect of a multi-path network that considers preserved localization and high-level semantics is explained in experiment 3.C.1).
\par Considering the trade-off between complexity and accuracy, MAP-Net is composed of three parallel paths for extracting multi-scale features as analyzed in experiment 3.C.2). The resolutions of feature maps are 1/4, 1/8 and 1/16 of the original image, with the corresponding channels are 64, 128 and 256 in each path.

\subsection{Attention-Based Feature Squeeze}

\begin{figure}[htbp]
\setlength{\abovecaptionskip}{-10pt}
\setlength{\belowcaptionskip}{-10pt}
\centering
\includegraphics[scale=0.19]{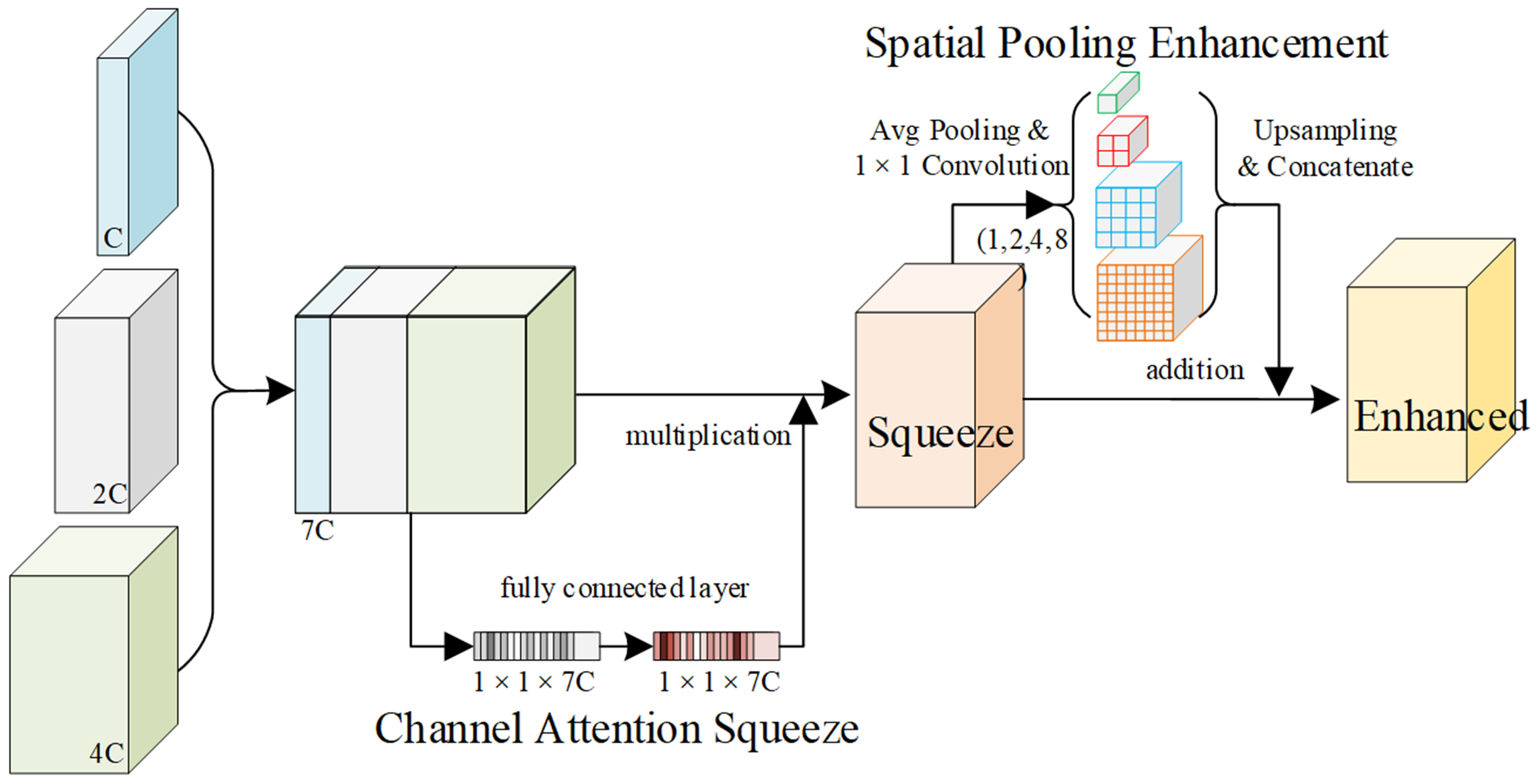} 
\caption{Feature semantic enhancement module. First, multi-scale features extracted from multi-paths are scaled by bilinear interpolation and concatenated. Then, the channel attention enhancement module adaptively squeezes significant channel-wise features to reconstruct optimal features. Finally, a spatial pooling enhancement module is introduced to capture global dependency for continuous building footprints.}
\label{Fig.f3_channel_enhance}
\end{figure}

\par Feature maps extracted from multi-paths have different dimensions. Higher-resolution features contain localization details and high-level semantic information, while lower resolution provides richer global features. The features are upsampled to 1/4 of the original image through bilinear interpolations and fused by concatenation, as shown in Fig. \ref{Fig.f3_channel_enhance}. It’s important to explore the optimal channel-wise combination of multi-scale features since they contain multi-level building localization details. We introduced a channel attention squeeze module adaptively measures the significance of each channel for optimizing the multi-scale features. A spatial pooling enhancement module is introduced to capture global dependence for continuously extracting building entities, especially for large buildings with low texture. The details are described as follows.
\par In previous CNN-based methods \cite{sun2019fusion,li2018multiple,ji2019scale}\cite{liu2019building,kang2019eu}, multi-scale features were concatenated directly for final pixel-wise prediction. Nevertheless, in our research, multi-scale features from different paths contain multi-level spatial localization and richer semantic representation. features with different scales have a dissimilar influence on building extraction. It’s the same as each channel, some of them may weaken the semantic representation but increase the computational complexity. In our research, multi-scale features from different paths contain spatial localization and richer semantic representation. It is necessary to distinguish valuable channel-wise features for accurately and efficiently extracting buildings, while a priori knowledge hardly weights the importance of each channel. The attention-based feature adaptive squeeze module inspired by \cite{hu2018squeeze} plays a role in learning the weight for each channel and automatically reconstructing the feature maps for optimal representation.
\par As illustrated in Fig. \ref{Fig.f4_basic}, a global average pooling operation produces a vector of length 7C from the concatenated multi-scale channel-wise feature, a fully connected layer with a weight parameter of 7C$\times$7C, followed by learning a weight vector with a length of 7C corresponding to each channel. The parameters of the fully connected layer are randomly initialized and gradually learned from the features. Finally, the vector that represents the significance of each channel is normalized by a sigmoid function and multiplied to the original features for reconstructing enhanced feature maps.
\par Due to the extracted features are always partially restricted by local receptive fields, a spatial pooling module is introduced to extract global dependence. The implementation is similar to \cite{zhao2017pyramid} except that the global features are generated by four average pooling layers with different sizes that are designed in accordance with the dimensions of features and added to the original feature maps pixel-wise for global spatial enhancement. It captures global relations spatially, which cannot be extracted from the CNN for the local respective field. Hence, extracted buildings have better integrity.

\subsection{Basic Block and Training Strategy}

\begin{figure*}[htbp]
\setlength{\abovecaptionskip}{-1pt}
\setlength{\belowcaptionskip}{-10pt}
\centering 
\subfigure[]{
\label{Fig.sub.1}
\includegraphics[scale=0.29]{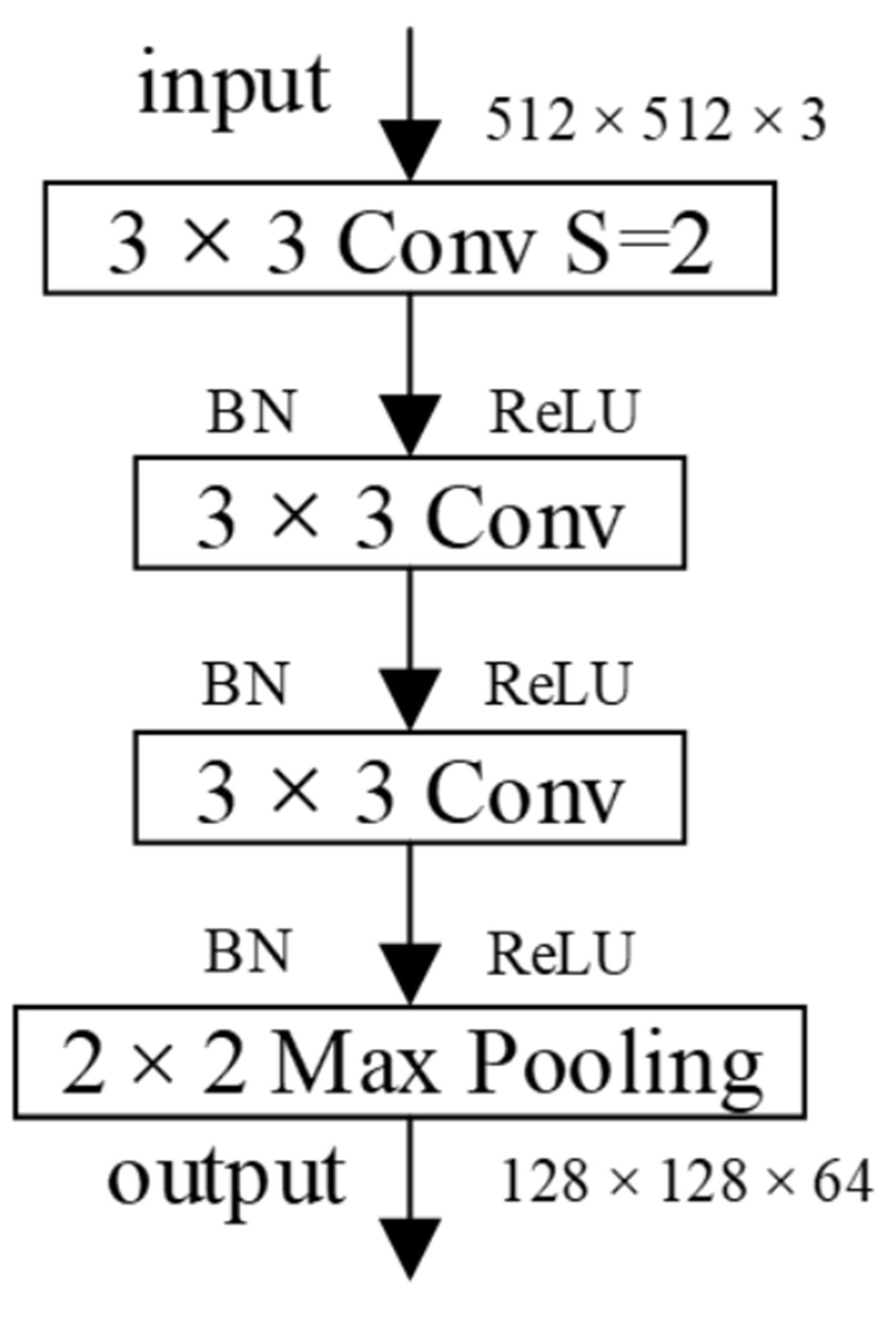}}
\subfigure[]{
\label{Fig.sub.2}
\includegraphics[scale=0.37]{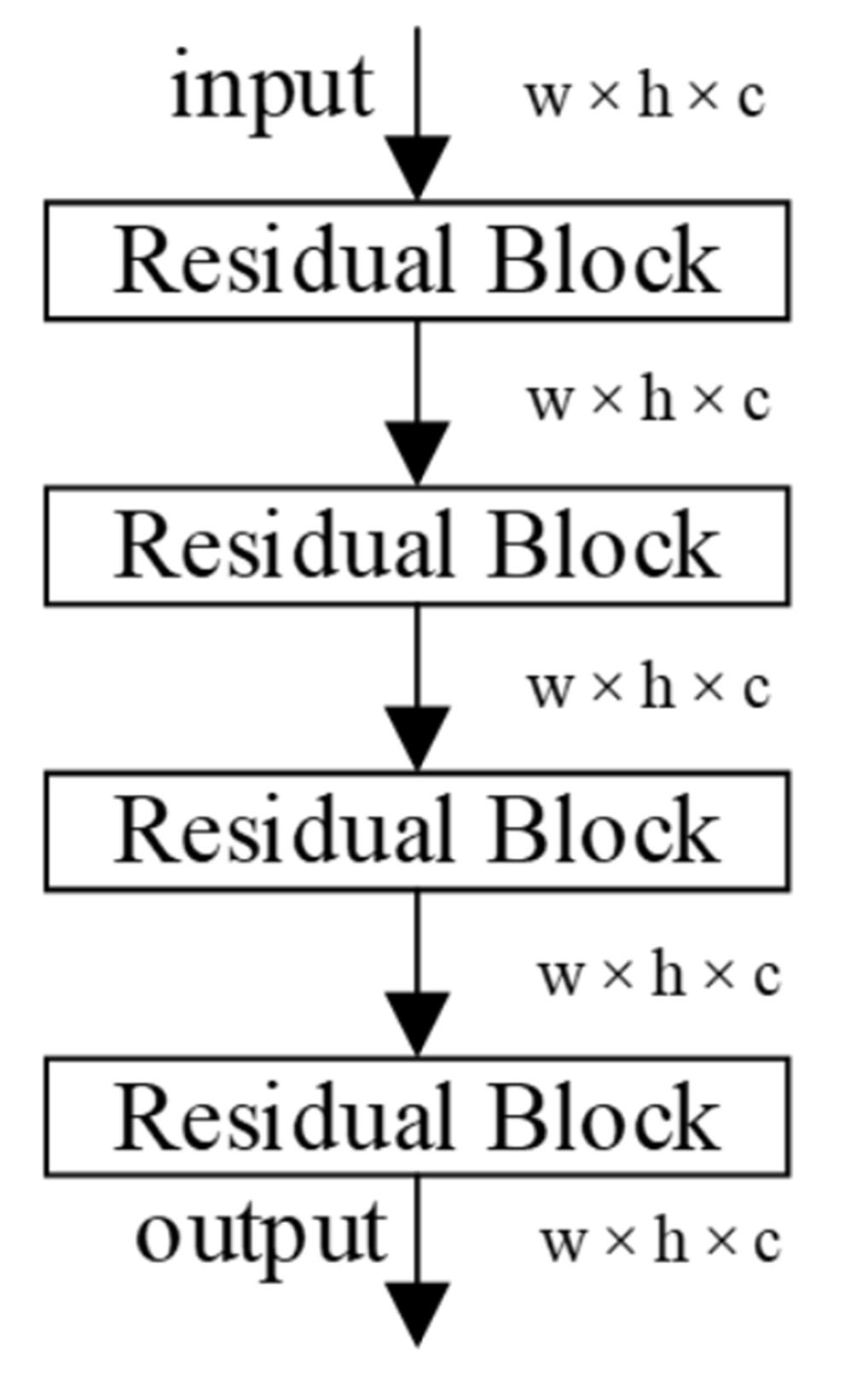}}
\subfigure[]{
\label{Fig.sub.3}
\includegraphics[scale=0.37]{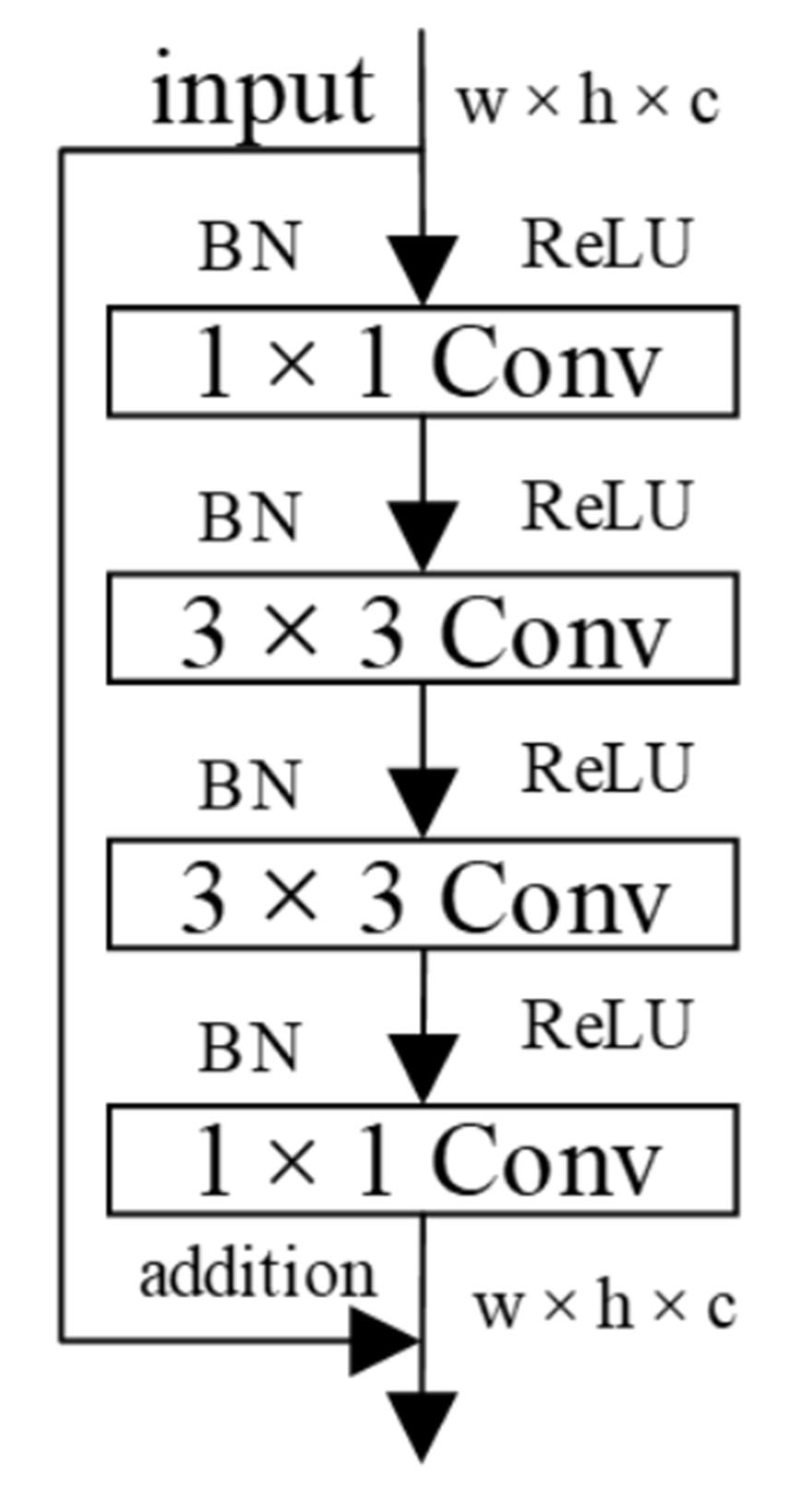}}
\subfigure[]{
\label{Fig.sub.4}
\includegraphics[scale=0.37]{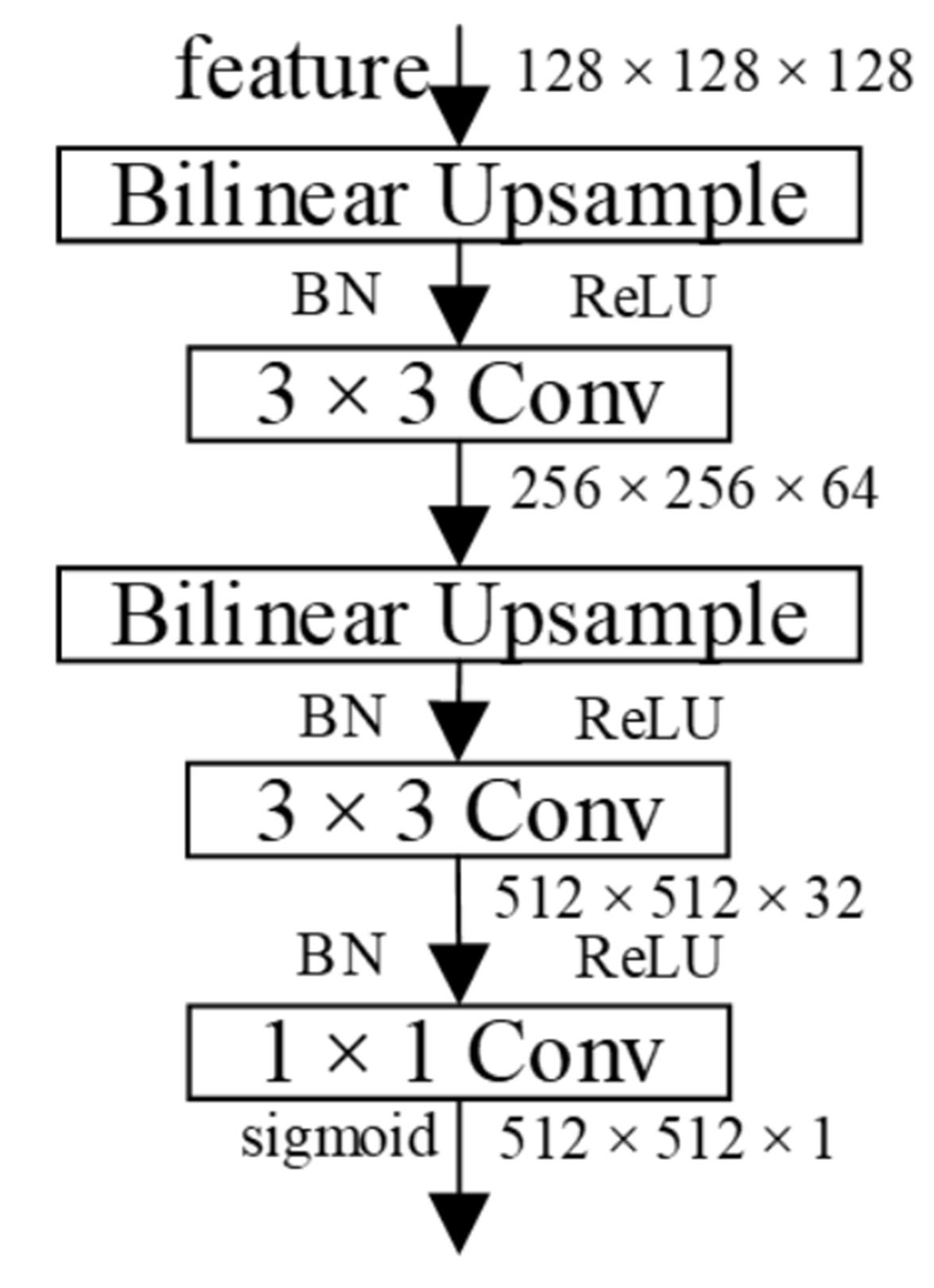}}
\caption{Detail of the basic blocks. (a) Downsample block. (b) Conv Block. (c) Residual block. (d) Upsample block.}
\label{Fig.f4_basic}
\end{figure*}

\par To avoid noise information contained in shallow features due to local receptive fields and decrease the computational complexity, a downsampling block is introduced to decrease the resolution of the features before the multi-path network, as shown in Fig. \ref{Fig.sub.1}. It consists of a stride convolution layer, two 3$\times$3 convolutional layers, and a max pooling layers to extract feature maps with 64 channels and 1/4 spatial resolution of the input image. Fig. \ref{Fig.sub.2} represents the conv block, which includes several residual blocks in series. The impact of different numbers of blocks on performance is explored in experiment 3.3.2. The residual block consists of a 1$\times$1 convolutional layer for reducing the dimensions of features, two 3$\times$3 layers for extracting features, and a 1$\times$1 convolutional layer for restoring dimensions to the input; a shortcut fuses input to output through element-wise addition, BN and ReLU execute before the convolutional layers, as illustrated in Fig. \ref{Fig.sub.3}. The building footprint extraction module is shown in Fig. \ref{Fig.sub.4}, the resolution of the features is recovered through bilinear interpolation in two stages, and the convolutional layers are used to decrease the number of channels. The output layer is a single-channel feature map with the same spatial resolution as input, and each value represents the probability that it belongs to building.
\par Our research was implemented in TensorFlow using a single 2080Ti GPU with 12 Gigabyte of memory. The Adam optimizer was chosen with an initial learning rate of 0.001, and beta1 and beta2 were set to default as recommended. All compared methods were trained from scratch for approximately 80 epochs until convergence and randomly rotated and flipped for data augmentation on three building datasets described in section III. The batch size was set to 4, given the restrictions of the GPU memory size and the same hyperparameters were maintained to compare the performance with different methods for equality.
\par Sigmiod cross-entropy loss was selected as the loss function because of the pixel-wise binary classification involved. The computation of loss at position $(i,j)$ is given as (\ref{1}); logits represent the predicted result and $y_{ij} \in \{0,1\}$  represents the ground truth; the sigmoid function was applied to logits to ensure that $p\in[0,1]$, as shown in (\ref{2}). The loss value is the average of $Loss_{ij}$ at all positions for an input image.

\begin{equation}\label{1}
Loss_{ij} = -[y_{ij}*\ln p_{ij} + (1-y_{ij})\ln (1-p_{ij})]
\end{equation}

\begin{equation}\label{2}
p_{ij} = Sigmoid(logits_{ij}) = \frac{1}{1+e^{-logits_{ij}}}
\end{equation}

\section{Experiment and Analysis}
\subsection{Dataset}
\par To evaluate the proposed method, we conducted a comprehensive experiment on three open datasets, including the WHU building dataset \cite{ji2018fully}, the Deep Globe Building Extraction Challenge dataset \cite{demir2018deepglobe} and the USSOCOM Urban 3D Challenge dataset \cite{goldberg2018urban}. The details are described as follows.
\par The WHU building dataset includes both aerial and satellite subsets with corresponding shapefiles and raster images. In our experiment, we selected the aerial subset, which has various appearances and scales of buildings, to evaluate the robustness of the proposed algorithm. It consists of more than 187,000 buildings, covering over a 450 $km^2$ area, with 30 $cm$ ground resolution. Each image has three bands, corresponding to red (R), green (G) and blue (B) wavelengths, with each image size of 512 $\times$ 512 pixels. There is a total of 8,188 tiles of images, including 4,736, 2,416 and 1,036 tiles as training, test and validation datasets, respectively. We conducted our experiment at its original provided dataset partitioning.
\par The Deep Globe Building Extraction Challenge dataset \cite{demir2018deepglobe} contains WorldView-3 satellite imagery captured from Las Vegas, Paris, Shanghai and Khartoum. In this research, the Las Vegas and Shanghai subsets were selected to evaluate the generalization performance of the proposed algorithm. There were approximately 243,382 buildings with 30 $cm$ ground resolution, covering over 1,216 $km^2$, and the size of each image was 650$\times$650 pixels. All images were randomly divided as 6:1:3 as the training set, validation set and test set,respectively.
\par The USSOCOM Urban 3D Challenge dataset \cite{goldberg2018urban} contains 208 orthorectified RGB, with corresponding DSM and digital terrain models (DTM) generated from commercial satellite imagery. It contains approximately 157,000 buildings, covering over 360 $km^2$ with a ground resolution of 50 $cm$, and the size of each image is 2048$\times$2048 pixels. DSM and DTM indicate the elevation of buildings, which obviously improves the building extraction performance. We used only the RGB images in our experiment to evaluate the performance of the proposed method. The training, validation and test set include 104, 62 and 42 tiles,respectively, as the original data partition method, and we randomly clipped the images to the size of 512$\times$512 pixels for training and testing.

\subsection{Evaluation Metric}
\par Generally, evaluation metric methodologies can be divided into two categories: pixel-level metrics and instance-level metrics. The pixel-level method counts the correctly classified and misclassified pixels pixel-wise. In the instance-level method, a building is correctly extracted only when the IoU between the prediction and ground truth is larger than a specific threshold. Semantic segmentation-based building footprint extraction aims to classify every pixel, whether or not it belongs to a building, for a specific input image. Therefore, we apply a pixel-level metric including precision, recall, F1-score and IoU to evaluate the performance of MAP-Net and other different methods.
\par There are four classifying conditions: true prediction on a positive sample (TP), false prediction on a positive sample (FP), true prediction on a negative sample (TN) and false prediction on a negative sample (FN). Precision represents the percentage of TP in total positive prediction, recall indicates the percentage of TP over the total positive samples, the F1-score is the weighted average of precision and recall, which considers both FP and FN, and IoU is the average value of the intersection of the prediction and ground truth over their union of the whole image set. Equations are given as follows:

\begin{equation}\label{3}
Precision = \frac{TP}{TP+FP}
\end{equation}

\begin{equation}\label{4}
Recall = \frac{TP}{TP+FN}
\end{equation}

\begin{equation}\label{5}
F1 = \frac{2*Precision*Recall}{Precision+Recall}
\end{equation}

\begin{equation}\label{6}
IoU=\frac{Precision*Recall}{Precision+Recall-Precision*Recall}
\end{equation}

\subsection{Experimental Setup}
\par In this section, we first analysed the significance of the proposed multi-path architecture for extracting multi-scale buildings with exact localization on boundaries compared with the popular encoder-decoder framework. Second, we explored the impact of different network parameters on the complexity and accuracy of MAP-Net on a specific dataset. Third, a contrast experiment was carried out to compare the performance of MAP-Net with four classic semantic segmentation algorithms on building extraction. Then, we compared the performance of most recent studies on building extraction based on the WHU dataset. Finally, we conducted an ablation experiment to validate the significance of the proposed network and analysed the trade-off between complexity and accuracy among the compared methodologies. Details are described in the following sections.
\subsubsection{Significance of Multi-path}

\begin{figure*}[h]
\setlength{\abovecaptionskip}{-3pt}
\centering
\includegraphics[scale=0.80]{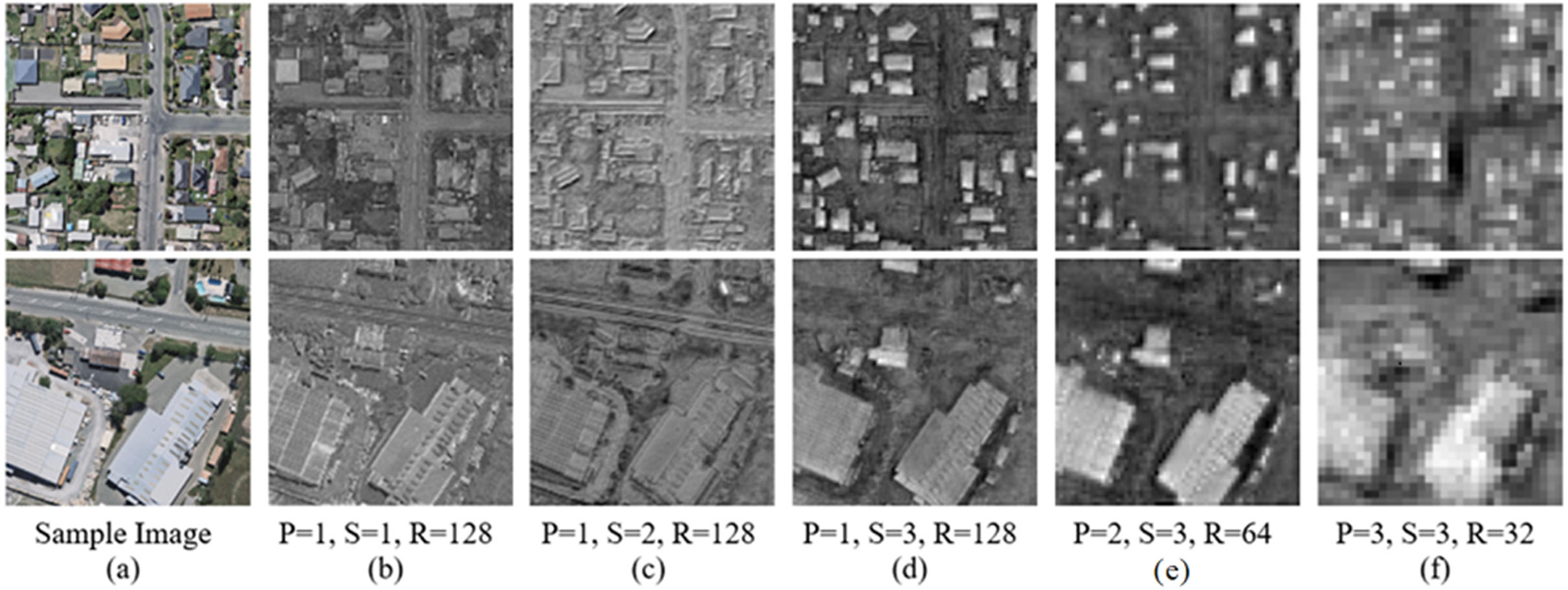} 
\caption{Feature maps extracted from localization-preserved multi-path networks with different paths (P), stages (S) and spatial resolutions (R) referred to in Fig. \ref{Fig.f1_main}. Column (a) represents two sample images containing multi-scale buildings. Columns (b-d) are feature maps extracted from path 1 with the same spatial resolution at the end of each stage. Columns (d-f) are the feature maps extracted from stage 3 but with decreasing spatial resolution in each path.}
\label{Fig.f5_feature_visual}
\end{figure*}

\par Feature maps extracted from the proposed localization-preserved multi-path network are visualized in Fig. \ref{Fig.f5_feature_visual}. Columns (b-d) are extracted from the path (P) 1 with the same spatial resolution (R) on each end of the stage (S) corresponding to the sample image in column (a). This indicates that feature maps with higher resolution extracted from deeper convolutional layers (larger S) retained richer semantics; in other words, building and background could be distinguished evidently. Columns (d-f) show the extracted feature maps from each path at the end of stage 3 with decreasing spatial resolution as shown in Fig. \ref{Fig.f1_main}. It shows that feature maps with lower resolution are more blurred at the edge of buildings, and in worse conditions, small buildings may be lost completely, as shown in column (f), due to the exact localization lost in downsampling operation during the feature extraction procedure.

\par Encoder-decoder-based networks fuse higher-resolution feature maps extracted from shallow layers, such as columns (b) or (c), to recover exact localization through a skip connection at the decoder stage, which introduces noise information for the coarse semantic features. In addition, small buildings may be lost in the lowest resolution feature maps, such as column (f), which cannot be refined accurately during the decoder stage. As a result, extracted building footprints were inaccurate on the boundary, or worse, small buildings were unrecognized.
\par Multi-path networks extract multi-scale feature maps through parallel paths. The resolutions of feature maps in each path were fixed, multi-scale features from the different paths were not fused in the whole feature extraction process. Features with higher spatial resolution preserved exact localization and contained rich semantic information through a deeper CNN, such as column (d). It is beneficial for extracting fine boundaries and small buildings compared to skip-connection with coarse shallow features as well as multi-scale features fusion  like HRNetv2. Additionally, the features with lower spatial resolution captured global semantic representations, which contribute to the extraction of large buildings. Multi-scale features extracted from multi-paths were combined and enhanced at the end of stage 3 to extract buildings with multi-scales, which makes up for the shortcoming of the existing network.

\subsubsection{Network Parameter}
\par The structure of the proposed network is mainly affected by the depth of the convolution network and the number of parallel paths on a specific dataset. Without losing generality, we designed an experiment to explore the potential impact of different structures on the performance of MAP-Net on the WHU dataset. Readers should be noticed that the network parameters will be different as dataset changing.
\par The depth is represented by the number of residual blocks (N-blocks) in each path, empirically, these were set from 3 to 6 in our experiments consider the trade-off between accuracy and complexity. Similarly, the number of paths (N-paths) was chosen from 2 to 4 according to the resolution of the input image. The IoU metric was used to evaluate the performance, and the number of trainable parameters (Para.) was counted to represent the complexity of the network.

\begin{figure}[htbp]
\setlength{\abovecaptionskip}{-0.28pt}
\centering
\includegraphics[scale=0.45]{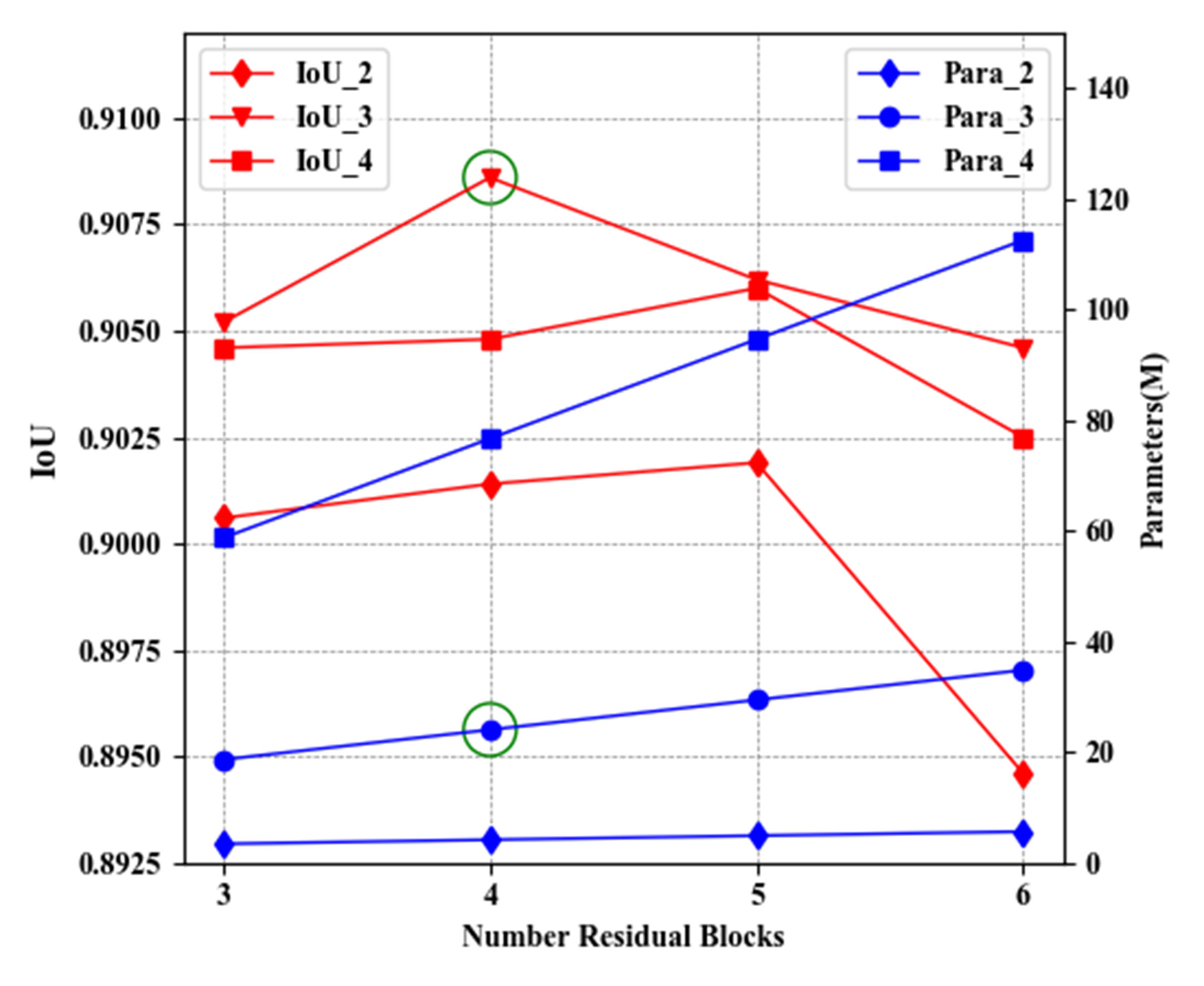} 

\caption{Performance of MAP-Net with different network structures. The diamond, circle and rectangle represent different numbers of paths. Red and blue represent the IoU score and trainable parameters (Para.), respectively. The horizontal axis indicates the number of residual blocks (N-blocks) in a convolutional block.}
\label{Fig.f6_commapre}
\end{figure}

\par The experimental result is illustrated in. Fig. \ref{Fig.f6_commapre} With the increase in N-blocks, the IoU score increased first, and then decreased after N-blocks were greater than a specific value, which may be explained by the complexity of the network growing with N-blocks increasing while weakening the generalization ability of the model. However, the Para. grows linearly with the increase in N-blocks, while increasing exponentially with the increase in N-paths since the generated path doubles the feature channels, which greatly increases the parameters during the feature extraction and enhancement stage.
\par Features with specific resolution were extracted from each path. The different number of paths impacts the combination of multi-scale semantic features fused in MAP-Net. When the N-paths were equal to 3, the IoU metric was better than that of 2 or 4, as shown by the solid line marked by the red circle in Fig. \ref{Fig.f6_commapre}.

\par Considering the balance between accuracy and complexity, the better structure of the MAP-Net was composed of three parallel paths, and each convolutional block consisted of four residual blocks, which contained fewer parameters and performed better than others on WHU dataset, as indicated by solid circles marked with green.
\par Besides, channels (C) of feature maps in each path and resolution of input images may also affect the performance of the proposed method. Generally, larger C will lead to better performance in the case of enough training data but increase complexity exponentially. Optimal paths of MAP-Net are probably affected by the resolution of remote sensing imagery for the different combinations of multi-scale features representation. The better structure of MAP-Net on another dataset could be explored similarly. As it’s not the focus of this paper, we will not state the detailed discussion here.

\begin{figure*}[hbp]
\setlength{\abovecaptionskip}{-13pt}
\centering
\hfil
\includegraphics[scale=0.391]{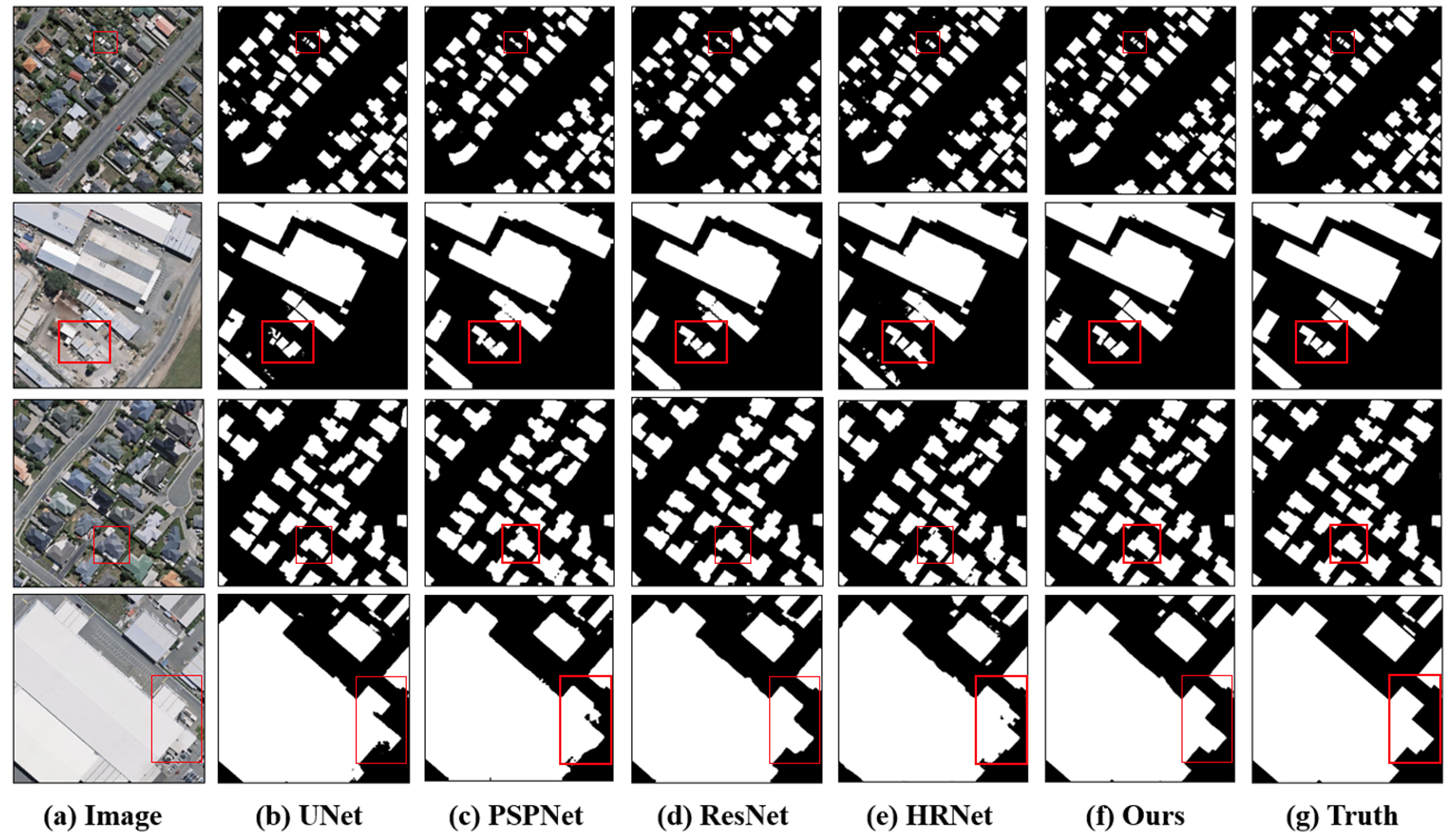} 
\caption{Example of results with the U-NetPlus, PSPNet, ResNet101, HRNetv2 and our proposed method on the WHU dataset. (a) Original image. (b) U-NetPlus. (c) PSPNet. (d) ResNet101. (e) HRNetv2. (f) Ours. (g) Ground truth.}
\label{Fig.f7_result_whu}
\end{figure*}

\begin{figure*}[hbp]
\setlength{\abovecaptionskip}{-5pt}
\setlength{\abovedisplayskip}{3pt}
\setlength{\belowdisplayskip}{3pt}
\centering
\hfil
\includegraphics[scale=0.391]{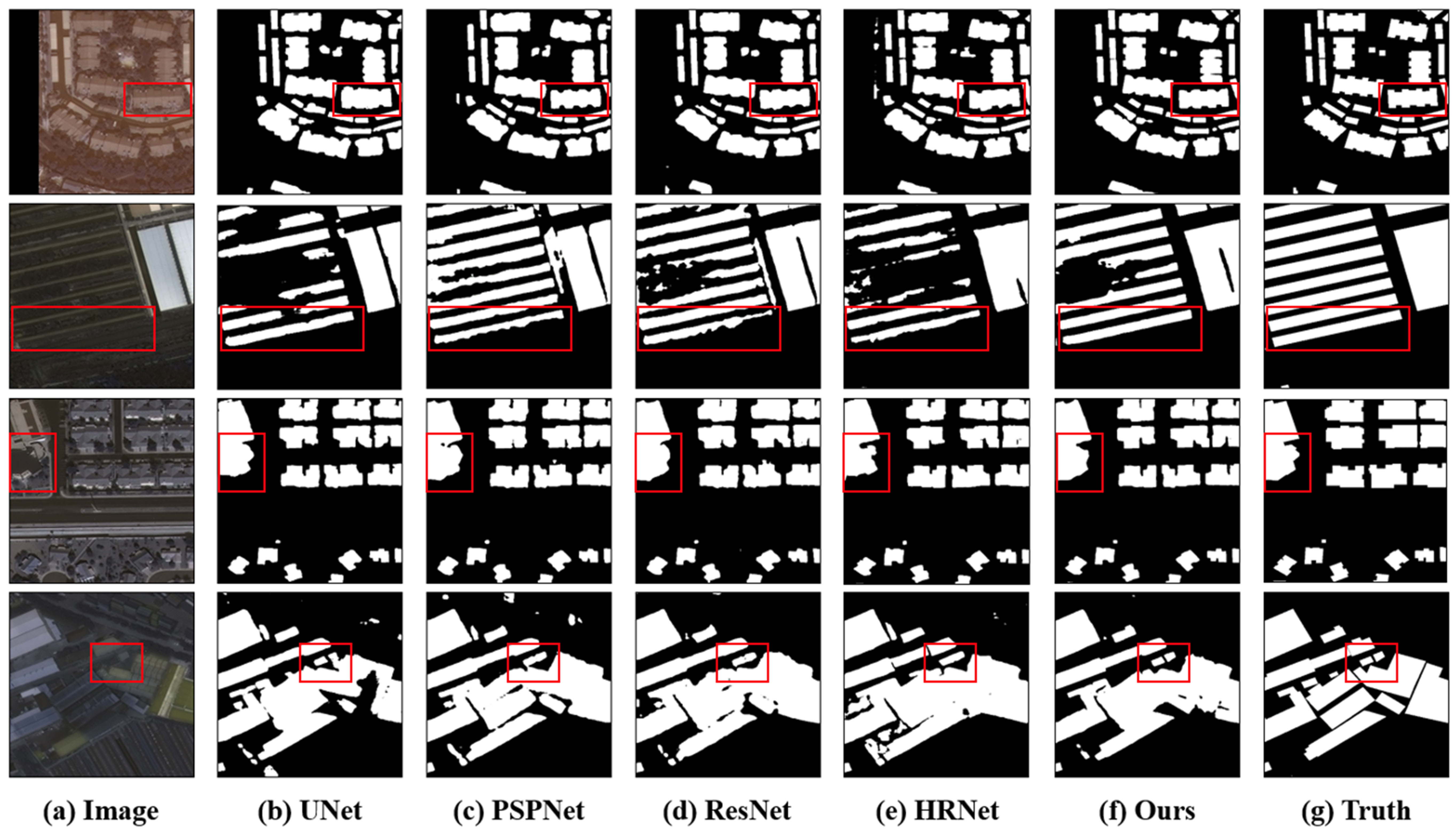} 
\caption{Example of results with the U-NetPlus, PSPNet, ResNet101, HRNetv2 and our proposed method on the Deep Globe dataset. (a) Original image. (b) U-NetPlus. (c) PSPNet. (d) ResNet101. (e) HRNetv2. (f) Ours. (g) Ground truth.}
\label{Fig.f8_result_spacnet}
\end{figure*}

\begin{figure*}[htbp]
\setlength{\abovecaptionskip}{-13pt}
\centering
\hfil
\includegraphics[scale=0.391]{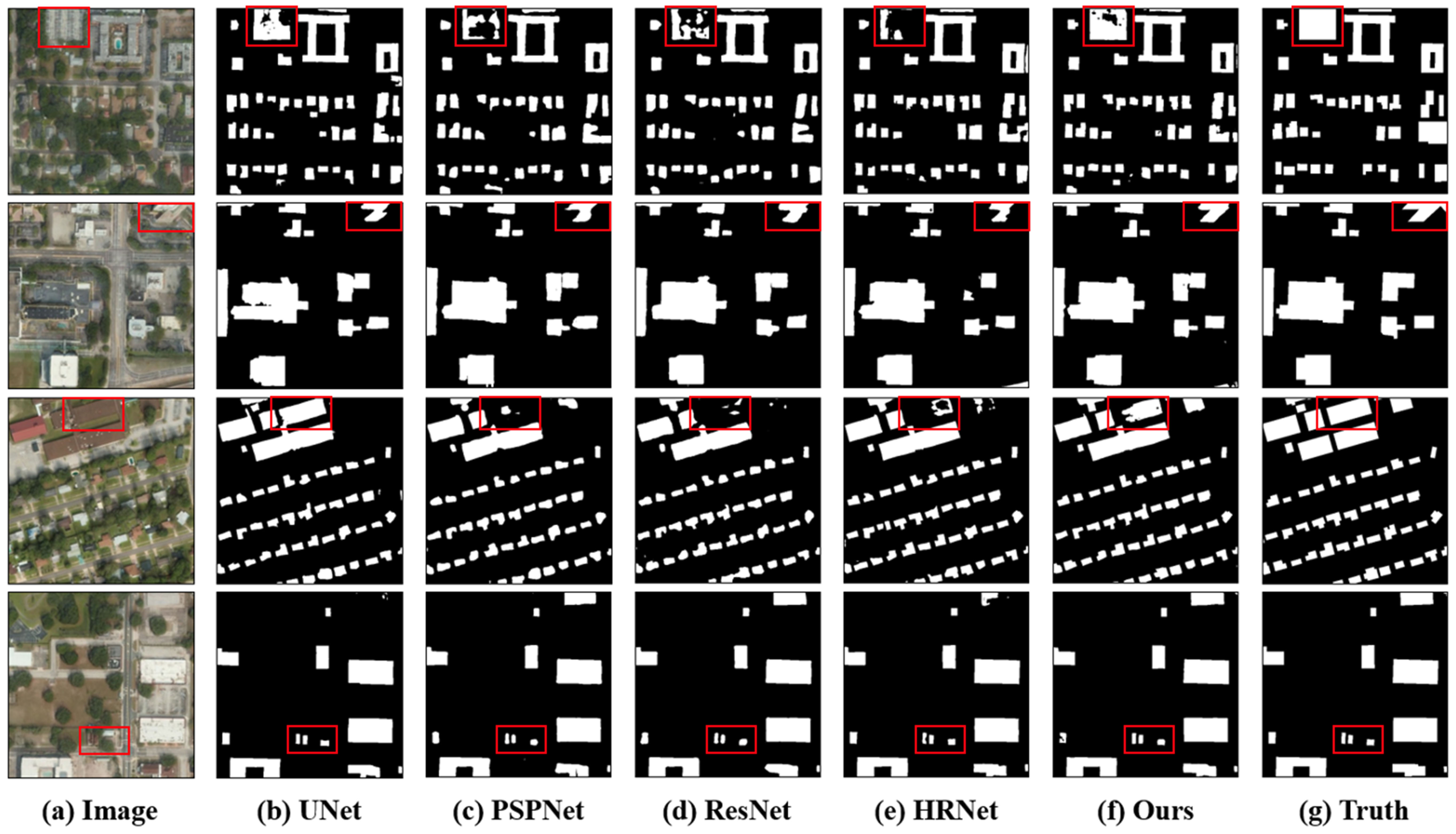} 
\caption{Example of results with the U-NetPlus, PSPNet, ResNet101, HRNetv2 and our proposed method on the Urban 3D dataset. (a) Original image. (b) U-NetPlus. (c) PSPNet. (d) ResNet101. (e) HRNetv2. (f) Ours. (g) Ground truth.}
\label{Fig.f9_result_urban3d}
\end{figure*}

\begin{table}[htbp]
\setlength{\abovecaptionskip}{-5pt}
\vspace{-0.3cm}  
\centering
\caption{Comparison of The State of The Art Methods and Ours on WHU Dataset.}
\label{table:1}
\normalsize
\setlength{\tabcolsep}{0.8mm}

\begin{center}
\begin{tabular}{@{}c|cccc@{}}
\toprule
Method    & IoU(\%)             & Precision(\%)       & Recall(\%)          & F1-score(\%)       \\ 
\midrule
U-NetPlus      & 88.75             & 94.85              & 93.25            & 94.04           \\
PSPNet    & 88.87             & 94.28              & 93.93            & 94.10           \\
ResNet101 & 89.18             & {94.47}            & 94.09            & 94.28           \\
HRNetv2   & {90.04}           & 94.16              & \textbf{95.37}   & {94.76}         \\
Ours      & \textbf{90.86}    & \textbf{95.62}     & {94.81}          &\textbf{95.21}   \\
\bottomrule
\end{tabular}
\end{center}
\end{table}

\begin{table}[htbp]
\setlength{\abovecaptionskip}{-5pt}
\vspace{-0.3cm}  
\centering
\caption{Comparison of The State of The Art Methods and Ours on Deep Globe Dataset. }
\label{table:2}
\normalsize
\setlength{\tabcolsep}{0.8mm}

\begin{center}

\begin{tabular}{@{}c|cccc@{}}
\toprule
Method    & IoU(\%)   & Precision(\%)    & Recall(\%)       & F1-score(\%)         \\
\midrule
U-NetPlus      & 76.34              & 89.07             & 84.23             & 86.59            \\
PSPNet    & 78.76              & 87.36             & \textbf{88.89}    & 88.12            \\
ResNet101 & {79.16}            & 89.10              & {87.65}           & 88.37            \\
HRNetv2   & 79.13              & {89.55}           & 87.17             & {88.35}          \\
Ours      & \textbf{80.63}     & \textbf{91.29}    & 87.35             & \textbf{89.28}   \\
\bottomrule
\end{tabular}
\end{center}
\end{table}

\begin{table}[htbp]
\setlength{\abovecaptionskip}{-5pt}
\vspace{-0.3cm}  
\centering
\caption{Comparison of The State of The Art Methods and Ours on Urban 3D Dataset.}
\label{table:3}
\normalsize
\setlength{\tabcolsep}{0.8mm}
\begin{center}
\begin{tabular}{@{}c|cccc@{}}
\toprule
Method    & IoU(\%)             & Precision(\%)       & Recall(\%)          &F1-score(\%)        \\
\midrule
U-NetPlus      & 84.56              & 92.59            & 90.69            & 91.63                        \\
PSPNet    & {86.19}            & {93.00}          & 92.17            & 92.46                        \\
ResNet101 & 86.17              & 92.83            & 92.31            & {92.57}                      \\
HRNetv2   & 86.15              & 92.74            & {92.38}          & 92.56                        \\
Ours      & \textbf{87.68}     & \textbf{93.42}   & \textbf{93.45}   & \textbf{93.44}               \\
\bottomrule

\end{tabular}
\end{center}
\end{table}

\subsubsection{Performance Evaluation}
\par To evaluate the performance of the proposed network, we conducted contrast experiments to compare MAP-Net with four SOTA methods, including UNet, PSPNet with ReNet50 backbone, ResNet101 and HRNetv2, on datasets \cite{demir2018deepglobe,goldberg2018urban,ji2018fully}. 
\par U-NetPlus achieved great improvement through re-designed encoder by VGG11 and replaced the transposed convolution with nearest-interpolation based on U-Net which widely used in remote sensing imagery segmentation. Since ResNet improved the training stability and performance of deeper CNN by introducing residual connection, we re-implemented PSPNet with ResNet50 for feature extraction and modified ResNet101 for building extraction through replacing the upsample module with the same as MAP-Net.
\par Experimental results are shown in TABLE \ref{table:1}, TABLE \ref{table:2} and TABLE \ref{table:3}. Our proposed method demonstrates a great improvement compared with other methods on three experimental datasets and obtains approximately 0.82\%, 1.50\% and 1.53\% IoU improvement and 0.45\%, 0.93\% and 0.88\% F1-score improvement on the WHU dataset, Deep Globe dataset and Urban 3D dataset, respectively, compared with the latest research HRNetv2. The best records are marked with bold.

\par To compare different methods, some example results on each dataset are presented in Fig. \ref{Fig.f7_result_whu}, Fig. \ref{Fig.f8_result_spacnet} and Fig. \ref{Fig.f9_result_urban3d}. Fig. \ref{Fig.f7_result_whu} shows extracted building footprints on the WHU dataset. There are four examples, including buildings with various appearances and scales. Columns (a) and (g) represent the original image and corresponding ground truth, and columns (b-f) are extracted results from U-NetPlus, PSPNet, ResNet101, HRNetv2 and MAP-Net, respectively.


\par The results show that our proposed method outperforms the other four compared methods obviously, especially by more accurately recognizing small buildings and more completely extracting large buildings, which benefits from the localization-preserved multi-path feature extraction network and the multi-scale feature adaptively enhancement module. The boundary of buildings is more exactly based on the ground truth.
\par Example results on the Deep Globe dataset and Urban 3D dataset are illustrated in Fig. \ref{Fig.f8_result_spacnet} and Fig. \ref{Fig.f9_result_urban3d}. Each column has the same meaning as presented in Fig. \ref{Fig.f7_result_whu}.

\subsubsection{Comparison of Recent Methods}
    \par We compare our methods to the most recent building extraction methods including CU-Net \cite{wu2018automatic}, SiU-Net \cite{ji2018fully}, SRI-Net \cite{liu2019building}, DE-Net \cite{liu2019net}, EU-Net \cite{liu2019net}, and MA-FCN \cite{wei2019toward}on WHU test Dataset to evaluate the performance of MAP-Net. TABLE \ref{table:4} shows the segmentation result of recent studies and ours, the best records are marked with bold. We didn’t reproduce their results since the source codes are not available.
 \par CU-Net \cite{wu2018automatic} introduced multi-constraint to enhance feature representation based on FCN. SiU-Net \cite{ji2018fully} designed multi-scale input with two weight-shared UNet which improve the result especially for a large building. DE-Net \cite{liu2019net} introduced lately segmentation techniques for building extraction based on Encoder-Decoder Network and achieved IoU accuracy higher than 90\%. Considering the significance of multi-scale features for building extraction, SRI-Net \cite{liu2019building} generated multi-level features through modified ResNet101 backbone and aggregate multi-scale contexts by spatial residual inception module, which achieved 89.23\% IoU on WHU dataset. EU-Net \cite{liu2019net} outperformed previous methods by a large margin with 90.56\% IoU since it captures multi-scale features by designed a deep spatial pyramids pooling (DSPP) and introduced focal loss to reduce the impact of mislabelling, which benefit for extracting fine details and multi-scale buildings. Most recent MA-FCN extract multi-scale features based on feature pyramid network considering the multi-scale building extraction and many post-processing strategies for boundary refinement. It’s better than EU-Net and achieved SOTA results which significantly benifit from multi-model voting strategy on the WHU dataset.
 \par As shown in Table IV, our study achieved 90.86\% in IoU metric and 95.21\% in F1-Score metric which outperform all the most recently studies. Especially, slightly outperforms MA-FCN on a result with limited performance improvement space without any pre-training and post-processing.
\par As shown in TABLE \ref{table:4}, our studies proposed a localization-preserved multi-path feature extraction network with a feature enhancement module for building extraction which achieved 90.86 IoU and outperformed the latest MA-FCN without pre-training and post-processing on WHU dataset.

\begin{table*}[htbp]
\setlength{\abovecaptionskip}{-5pt}
\centering
\caption{Comparison of Most Resent Building Extraction Methods on WHU dataset.}
\label{table:4}
\normalsize
\setlength{\tabcolsep}{5.5mm}
\begin{center}
\begin{tabular}{@{}c|cccc@{}}
\toprule
Method                          & IoU(\%)             & Precision(\%)       & Recall(\%)          &F1-score(\%)      \\
\midrule
CU-Net \cite{wu2018automatic}    & 87.10              & 94.60            & 91.70            & 93.13                   \\
SiU-Net \cite{ji2018fully}     & 88.40              & 93.80            & 93.90            & 93.85                      \\
SRI-Net \cite{liu2019building} & 89.23              & \textbf{95.67}            & 93.69            & 94.51                     \\
DE-Net \cite{liu2019net}     & 90.12$\pm$ 0.24      & 95.00$\pm$ 0.16     & 94.60$\pm$ 0.19     & 94.80$\pm$ 0.18                   \\
EU-Net \cite{kang2019eu}     & 90.56              & 94.98            & \textbf{95.10}          & 95.04                         \\
MA-FCN \cite{wei2019toward}   & 90.70              & 95.20            & \textbf{95.10}          & 95.15                         \\
MAP-Net (Ours)      & \textbf{90.86}     & {95.62}   & {94.81}   & \textbf{95.21}                         \\
\bottomrule

\end{tabular}
\end{center}
\end{table*}

\subsubsection{Ablation Experiments}

\par To explore the contributions of different modules of the MAP-Net, we conducted ablation experiments on the WHU dataset and evaluate the accuracy of IoU, precision, recall, and F1-score.
\par Firstly, we validated the performance of the localization-preserved strategy (Baseline) compared with HRNetv2 and HRNetv1. Secondly, we explore the effect of feature squeeze module and global enhancement module on MAP-Net. Finally, we compare the influence of shallow features (F) which have a half resolution of the input imagery on MAP-Net with skip-connection. 
\par Experimental results are recorded in TABLE \ref{table:table5}, the highest values are highlighted in bold and the visualization of extracted results from each method shown in Fig. \ref{Fig.f10_visual} for the comparison of different methods. The false prediction marked with red, the false-positive marked with blue and the true positive marked with green.
\par Our baseline based on HRNetv2 and optimized the architecture on multi-scale features fusion as described in Section II.B, the baseline methods outperform the HRNetv2 by 0.24\% on the IoU since modified multi-path localization-preserved strategy without feature fusion in the encoder, and higher 0.74\% than the HRNetv1 on the IoU for fusing multi-scale features on the decoder the same as HRNetv2.
\par Based on baseline, the channel-wise feature adaptively squeezes module based on the attention mechanism called (C) and the spatial pooling enhancement module named (S) improved the IoU performance by 0.29\% and 0.34\% respectively. The MAP-Net achieved 0.58\% IoU arise with feature squeeze module and global enhancement module compared to our baseline. As described in Fig. \ref{Fig.f1_main}, the resolution of fused features is a quarter of input image in our MAP-Net. To evaluate the influence of shallow feature which has a higher resolution on building extraction, we concatenate the feature with half-resolution of input image extracted from the downsample block to the upsample block thought skip-connection named MAP-Net+F.
\par According to the experimental results, adding shallow features improved the accuracy by 0.04\%, but the introduced shallow features also increase the coarse noise information, as shown in the last column of Fig. \ref{Fig.f10_visual}, resulting in inaccurate building boundary. The MAP-Net can get a smoother edge of the building without introducing shallow features, and reduce unnecessary error recognition with little accuracy loss.
\par It is worth noting that our algorithms from No.3 to No.6 obtained higher precision measures than HRNetv2 with the same threshold equals to 0.5. A probable explanation is that our methods suppressed false positive prediction, which contributed to accurate multi-scale features extracted from localization-preserved multi-path networks, without fusion during feature extraction. The same conclusion can be inferred from other datasets according to TABLE \ref{table:2} and TABLE \ref{table:3}.

\begin{table*}[htbp]
\setlength{\abovecaptionskip}{-3pt}
\centering
\caption{INFLUENCE OF DIFFERENT MODULES. (M): MULTI-PATH EXTRACTION.
(C): CHANNEL FEATURE SQUEEZE. (S): SPATIAL FEATURE ENHANCE. (F) CONNECTION WITH SHALLOW FEATURES}
\label{table:table5}
\normalsize
\setlength{\tabcolsep}{5.5mm}
\begin{center}
\begin{tabular}{@{}cc|cccc@{}}
\toprule
No. & Method & IoU(\%) & Precision(\%) & Recall(\%) & F1-Score (\%) \\
\midrule
1    & HRNetv1         & 89.54            & 94.55            & 94.41               & 94.48              \\
2    & HRNetv2         & 90.04            & 94.16            & \textbf{95.37}      & 94.76                  \\
3    & Baseline        & 90.28            & 95.03            & 94.76               & 94.89                 \\
4    & Baseline+ C    & 90.57            & 95.44            & 94.67               & 95.05                    \\
5    & Baseline+ S    & 90.62            & 95.21            & 94.95               & 95.08               \\
6    & MAP-Net(Ours)     & {90.86}   & \textbf{95.62}   & 94.81               & {95.21}         \\ 
7    & MAP-Net + F       & \textbf{90.90}   & {95.18}   & 95.29               & \textbf{95.23}         \\ 
\bottomrule
\end{tabular}
\end{center}
\end{table*}

\begin{figure*}[htbp]
\setlength{\abovecaptionskip}{-5pt}
\vspace{-0.3cm}  
\centering
\hfil
\includegraphics[scale=0.30]{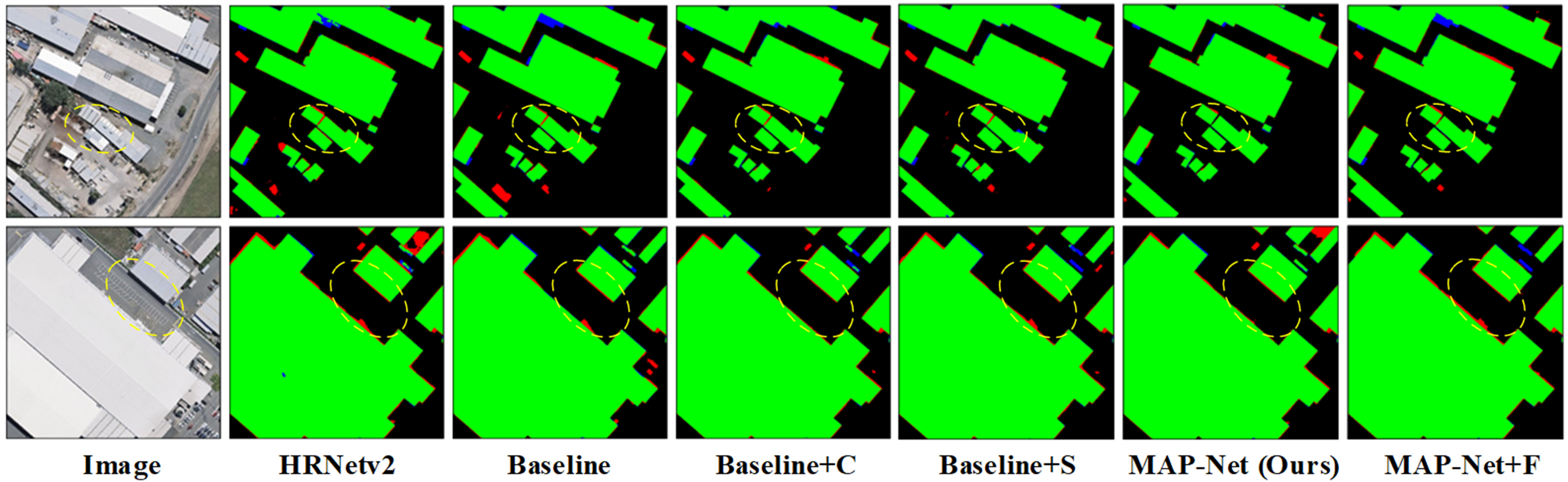} 
\caption{Visualization of extracted results on ablation experiments. False prediction marked with red, false positive marked with blue and true positive marked with green.}
\label{Fig.f10_visual}
\end{figure*}

\subsubsection{Complexity of MAP-Net} 

\par Our proposed algorithm extracted features with multi-scales; specifically, some paths needed to process feature maps with large resolution to preserve exact localization in the whole network, which could lead to large numbers of parameters. To validate the trade-off between the performance and complexity of MAP-Net, we compared FLOPs, trainable parameter and IoU score of related methods on the WHU dataset.

\par As shown in TABLE \ref{table:6}, The U-NetPlus has the lowest complexity but with poor performance. ResNet101 is the most complicated model due to the most numbers of convolutional layers and the highest numbers of channels among the related methods. HRNetv2 has slightly more parameters than HRNetv1 and performs better since the concatenate fusion of multi-scale features in the decoder stage. 
\par The complexity of our baseline has reduced with higher performance compared to HRNetv2 since the re-design structure of encoder. Although the MAP-Net has increased complexity after introduced a feature enhancement module, the performance has greatly improved with far fewer FLOPs than HRNetv2. 
\par Although MAP-Net maintains a high-resolution feature map in the whole feature extraction process, which may lead to a large number of parameters, the number of channels remains small, allowing it to efficiently extract multi-scale features. To compare the complexity and performance among related methods more intuitively, we presented the experimental results in Fig. \ref{Fig.f11_efficient_compare}. The number of trainable parameters and IoU precision represents complexity and performance for each related method. The radius of the green circle indicates the size of the model file. Our proposed methods maintain higher accuracy and lower complexity compared with other related methods. 

\begin{table}[htbp]
\setlength{\abovecaptionskip}{-5pt}
\setlength{\belowdisplayskip}{10pt}
\centering
\caption{Comparison of Related Methods on FLOPs, Trainable Parameters and IoU Score on WHU Test Dateset.}
\label{table:6}
\normalsize
\setlength{\tabcolsep}{1.2mm}
\begin{center}
\begin{tabular}{@{}c|cccc@{}}
\toprule
Method    & FLOPs(M)             & Parameters(M)       & IoU(\%)              \\
\midrule
U-NetPlus & 17.28            & 8.636            & 89.51                        \\
PSPNet    & 93.48            & 46.73            & 88.87                         \\
ResNet101 & 104.31           & 52.11            & 89.18                          \\
HRNetv1   & 57.18              & 28.54            & 89.54                         \\
HRNetv2   & 59.20              & 29.54            & 90.04                          \\
Baseline  & 46.78              & 23.35            & 90.28                       \\
Baseline + C  & 47.16           & 23.54             & 90.63                       \\
Baseline + S  & 47.58           & 23.75             & 90.56                        \\
MAP-Net (Ours)   & 48.09        & 24.00             & 90.86                         \\
\bottomrule

\end{tabular}
\end{center}
\end{table}

\begin{figure}[htbp]
\setlength{\abovecaptionskip}{-15pt}
\vspace{-0.3cm}  
\centering
\includegraphics[scale=0.5]{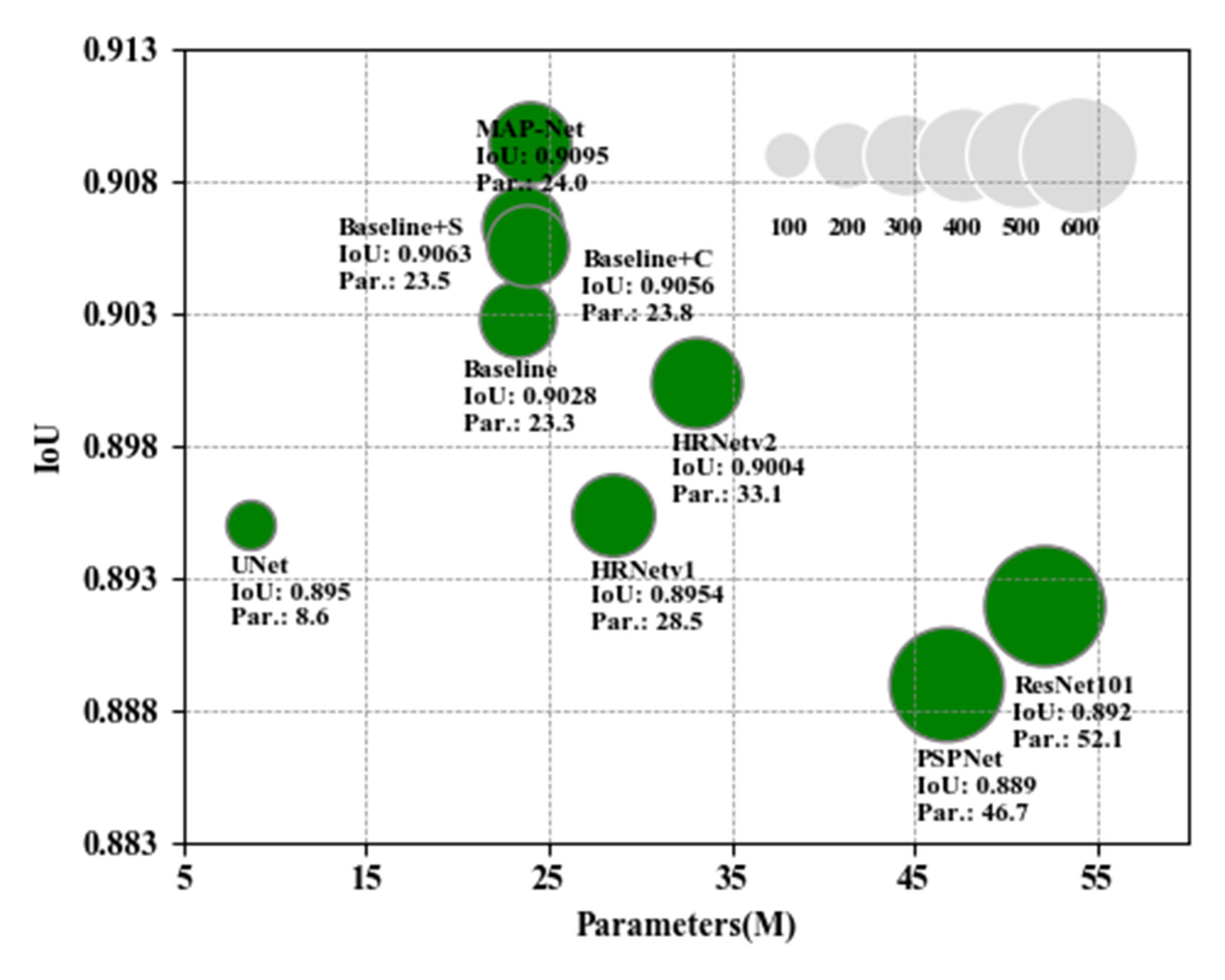} 
\caption{Complexity and accuracy comparison among related methods. IoU precision and the number of trainable paramet
ers for each method are marked in the figures. The radius of the circle indicates the size of the model file.}
\label{Fig.f11_efficient_compare}
\end{figure}



\section{Discussion and Conclusion}

\par To solve the problem of extracted building footprints with inaccurate boundaries and possibly lost small buildings as well as discontinuous for large-scale buildings, in this research, we proposed a novel localization-preserved multi-path feature extraction network inspired by HRNetv2 with a adaptively  multi-scale feature optimal fusion module and spatial enhancement module for building footprint extraction. Multi-scale features extracted from parallel multi-paths that contain multi-level local details, as well as rich semantic representations allow it to accurately extract building footprints with exact edges and recognize small buildings. The enhanced module further reconstructs and optimizes features in channel and spatial aspects, which suppresses the holes and extracts continuous footprints for large buildings.
\par The experiments on three different benchmarks demonstrate that the MAP-Net outperforms other classic semantic segmentation algorithms with higher accuracy and lower complexity and achieved SOTA result on WHU dataset among the most recent building extraction methods. In addition, we conducted an ablation experiment to evaluate the significance of the proposed module and proved that localization-preserved multi-path network extraction of buildings achieves higher precision than previous methods.
\par Generally, our research provides a new approach for accurately and efficiently extracting multi-scale objects that are common in the real world. Currently, our experiments are implemented in building extraction, and we will further study multiclass extraction tasks, such as land cover, to achieve automatic interpretation of remote sensed imagery in future work.


%




\ifCLASSOPTIONcaptionsoff
  \newpage
\fi



%
\bibliographystyle{ieeetr}
\bibliography{pic/mapnet_refrence}
%

\vfill

\end{document}